\documentclass[11pt,letterpaper]{iffyuan}

\usepackage[all]{hypcap}
\usepackage{algorithm}
\usepackage{algorithmicx}
\usepackage{algpseudocode}
\usepackage{subcaption}
\expandafter\def\csname ver@subfig.sty\endcsname{}
\usepackage{subfig}
\usepackage{float}
\usepackage{bigstrut}
\usepackage{mathtools}
\usepackage{amsthm}
\usepackage{nicefrac}
\usepackage{bxcoloremoji}
\usepackage{fdsymbol}
\usepackage{wrapfig}
\usepackage{lipsum}
\usepackage{stackengine}
\usepackage{enumitem}
\usepackage{adjustbox}
\usepackage{rotating}
\usepackage{makecell}
\usepackage{xspace}
\usepackage{tikz}
\usepackage{array}
\usepackage{ulem}
\tcbuselibrary{skins}

\newcommand{\method}{RoboPIN}
\newcommand{\pincot}{PinCoT}
\newcommand{\teacher}{Qwen3-VL-32B-Instruct}
\newcommand{\student}{Qwen3-VL-4B-Instruct}
\newcommand{\reasoninganchor}{reasoning anchor}
\newcommand{\dataset}{PIN-170K}
\newcommand{\codeurl}{https://github.com/quark404/RoboPIN}
\newcommand{\dataseturl}{https://huggingface.co/datasets/QwQ2/RoboPIN-Datasets}

\definecolor{lightblue}{rgb}{0.22,0.45,0.70}%
\definecolor{Gray}{gray}{0.95}
\definecolor{Cornsilk}{rgb}{1.0, 0.97, 0.86}

\title{RoboPIN: Grounded Embodied Reasoning via Pinned Chain-of-Thought}
\runningtitle{RoboPIN: Grounded Embodied Reasoning via Pinned Chain-of-Thought}

\author[1,*]{Yaoting Huang}
\author[1,*,\dag]{Yifu Yuan}
\author[1]{Linqi Han}
\author[1]{Chengwen Li}
\author[1]{Shuoheng Zhang}
\author[1]{Xianze Yao}
\author[1]{Hongyao Tang}
\author[1]{Yan Zheng}
\author[1,\dag]{Jianye Hao}

\affiliation[1]{Tianjin University}
\contribution{* These authors contributed equally to this work. \\
\dag \hspace{.2em} Corresponding authors: Yifu Yuan (\href{mailto:yuanyf@tju.edu.cn}{yuanyf@tju.edu.cn}), Jianye Hao (\href{mailto:jianye.hao@tju.edu.cn}{jianye.hao@tju.edu.cn})}

\abstract{%
Embodied reasoning requires models to perceive task-relevant objects and spaces in physical environments and maintain consistent visual grounding throughout multi-step reasoning. However, current vision-language models rely on text-only or coordinate-augmented chain-of-thought, where entity references remain implicit and ambiguous. This may cause the reasoning process to decouple from visual evidence, entity references to drift across steps, and a causal disconnection between the reasoning trajectory and the final answer, with these problems further amplified in multi-view scenarios due to cross-view appearance changes. To address these issues, we propose Pinned Chain-of-Thought (\pincot{}), a structured reasoning paradigm that pins every reasoning step to visual evidence. \pincot{} introduces the concept of \reasoninganchor{}, which binds each task-relevant entity to a structured visual anchor with entity name, unique identity, view index, and spatial grounding, enabling consistent entity tracking across reasoning steps and views. We build a fully automated data generation pipeline to construct \dataset{}, a high-quality \pincot{}-formatted reasoning dataset. We then train \method{} through three-stage post-training that progressively injects embodied knowledge, structured reasoning ability, and process-supervised alignment, with rewards that directly constrain both anchor localization and identity consistency during reasoning. On 14 benchmarks covering embodied spatial reasoning, multi-view reasoning, and pointing, \method{} with only 4B parameters surpasses 7B level open-source embodied models on average, achieving a 12\% average improvement over the strongest 7B baseline, Mimo-Embodied. Further analysis shows that \pincot{} improves grounding accuracy and cross-step identity consistency, validating the effectiveness of process supervision.
}

\metadata[Keywords]{Embodied Reasoning, Spatial Reasoning, Multi-view Reasoning}
\metadata[\raisebox{-1.5pt}{\includegraphics[height=1.03em]{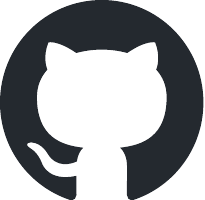}}\ Code]{\href{\codeurl}{\codeurl}}
\metadata[\raisebox{-1.5pt}{\includegraphics[height=0.96em]{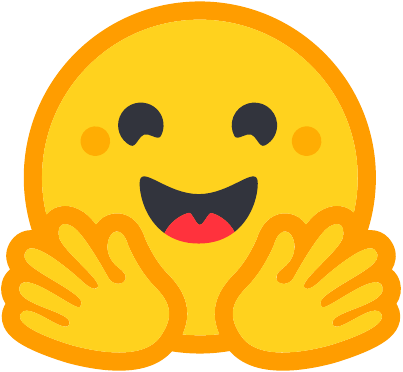}}\ Dataset]{\href{\dataseturl}{\dataseturl}}
\metadata[\faEnvelope\ Contact]{\href{mailto:quark404@tju.edu.cn}{quark404@tju.edu.cn}}

\setteamlogo{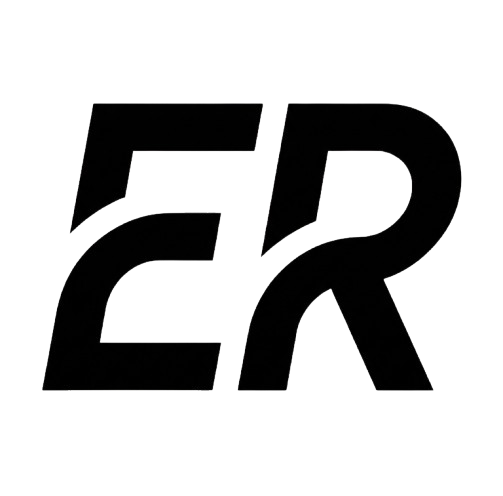}

\begin{document}

\maketitle

\section{Introduction}
Embodied reasoning is a core capability for intelligent systems operating in the physical world~\cite{sermanet2024robovqa,chen2026egoplan,team2025gemini}. Tasks such as spatial reasoning, fine-grained pointing, and long-horizon planning require models to accurately perceive task-relevant objects and spatial regions in complex environments, while maintaining tight coupling between reasoning steps and visual evidence throughout multi-step inference~\cite{chen2024spatialvlm,cheng2024spatialrgpt,yuan2024robopoint,cheng2025pointarena,song2025robospatial}.

Despite the necessity of this tight coupling, current Vision-Language Models (VLMs) struggle to maintain it during embodied tasks.~\cite{du2024embspatial,song2025robospatial,fu2024blink}. Although chain-of-thought reasoning has advanced beyond pure text into multimodal and visually grounded forms~\cite{zhang2023multimodal}, existing methods still fall short on reasoning consistency. Multimodal chain-of-thought methods~\cite{mitra2024compositional,zheng2023ddcot,shao2024visual} incorporate visual information into the reasoning process, but reasoning steps remain largely textual and lack explicit spatial grounding. Coordinate-augmented approaches~\cite{chen2023shikra,li2025vocot,liao2024reasoning} go further by embedding coordinates or bounding boxes into reasoning chains, yet these spatial references are often isolated and lack explicit identity tracking across steps. As illustrated in Figure~\ref{fig:teaser}, this leads to two core problems. First, \textbf{reasoning decouples from visual evidence}: entity references within the reasoning process are implicit textual descriptions that cannot guarantee each step looks at the correct target, and cross-step entity references are prone to drift, especially when multiple instances of the same category co-exist in the scene. Second, \textbf{reasoning-to-answer misalignment}: when intermediate steps suffer from reference drift, the final answer loses its reliable visual grounding. Consequently, a model might correctly analyze object~A initially, but silently shift its attention and point to object~B in the final prediction. Current paradigms fail to detect this disconnection because the underlying causal chain is broken. These issues are further amplified in multi-view settings, these problems are further amplified because the same object undergoes appearance changes across viewpoints~\cite{wang2025towards,feng2025seeing,yang2025mmsi}. We unify these issues as the \textit{reasoning consistency} problem in embodied reasoning, spanning cross-step and cross-view consistency. The root cause is the lack of a unified \textit{\reasoninganchor{}} mechanism to persistently bind visual evidence throughout reasoning.

\begin{figure*}[t]
  \centering
  \includegraphics[width=0.98\linewidth]{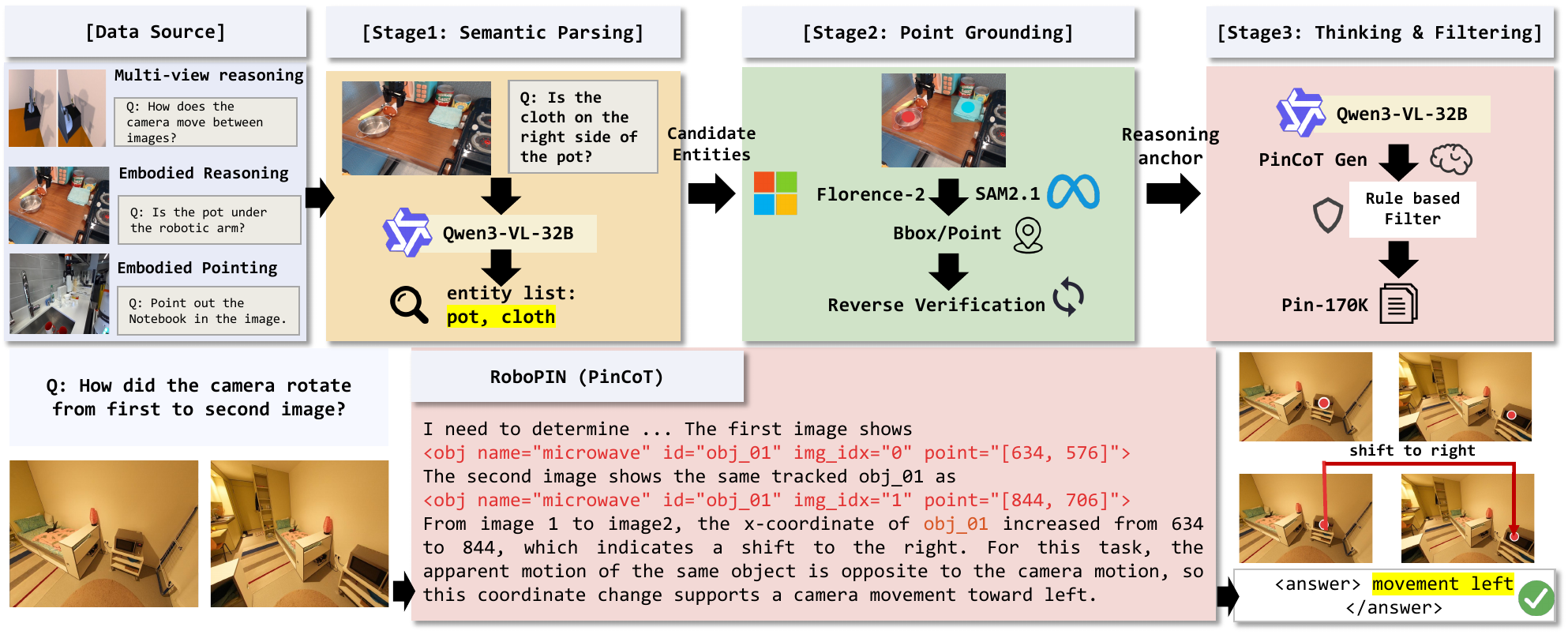}
  \vspace{-5pt}
  \caption{Overview of \method{}. Our framework enables grounded embodied reasoning through \pincot{}, which pins every reasoning step onto visual evidence with identity-tracked anchors.}
  \label{fig:intro_overview}
\end{figure*}

To address these issues, we propose \textit{Pinned Chain-of-Thought} (\textit{\pincot{}}), a structured reasoning paradigm that ``pins'' every reasoning step onto visual evidence. The core of \pincot{} is the \textit{\reasoninganchor{}}: each task-relevant entity is bound to a structured visual anchor with a name, unique identity, view index, and spatial grounding, rather than a one-shot perceptual detection result. During reasoning, an entity is introduced with a full tag establishing its localization and identity upon first appearance; subsequent steps refer to the same entity via its ID, forming a verifiable chain. enables identity-consistent reasoning, which extends to multi-view settings where the same object shares its ID across different viewpoints. Unlike coordinate-embedding methods such as Shikra~\cite{chen2023shikra}, region-grounding approaches such as Ferret~\cite{you2023ferret}, and visual chain-of-thought methods such as VoCoT~\cite{li2025vocot}, \pincot{} does not perform static grounding external to the reasoning process, but rather treats these visual references as persistently trackable anchors within the reasoning chain. To obtain supervision in the \pincot{} format~\cite{shao2024visual,zawalski2024robotic}, we build a fully automated pipeline and use it to construct the \dataset{} reasoning dataset, as shown in Figure~\ref{fig:intro_overview}. The pipeline transforms multimodal data into \pincot{}-format reasoning supervision through three stages: semantic parsing, precise point grounding, and reasoning chain generation, incorporating dedicated quality assurance throughout the process.

Building on this framework and data, we train \method{} through SFT, CoT-SFT, and RFT three-stage post-training, which respectively inject embodied knowledge, structured reasoning ability, and process-supervised alignment~\cite{yuan2025embodied}. The RFT stage employs a composite reward function that simultaneously supervises the reasoning process and the final answer~\cite{lightman2023let}, effectively mitigating reference drift across reasoning steps.

Extensive experiments show that \method{} achieves best average performance across 14 benchmarks covering embodied spatial reasoning, multi-view reasoning, and pointing tasks~\cite{du2024embspatial,song2025robospatial,ray2024sat,wang2025towards,feng2025seeing,fu2024blink}. Further analysis confirms that \pincot{} substantially improves the localization accuracy of reasoning anchors and cross-step identity consistency, validating the effectiveness of process supervision on reasoning quality.
The main contributions are as follows:

\begin{figure*}[t]
    \centering
    \includegraphics[width=\textwidth]{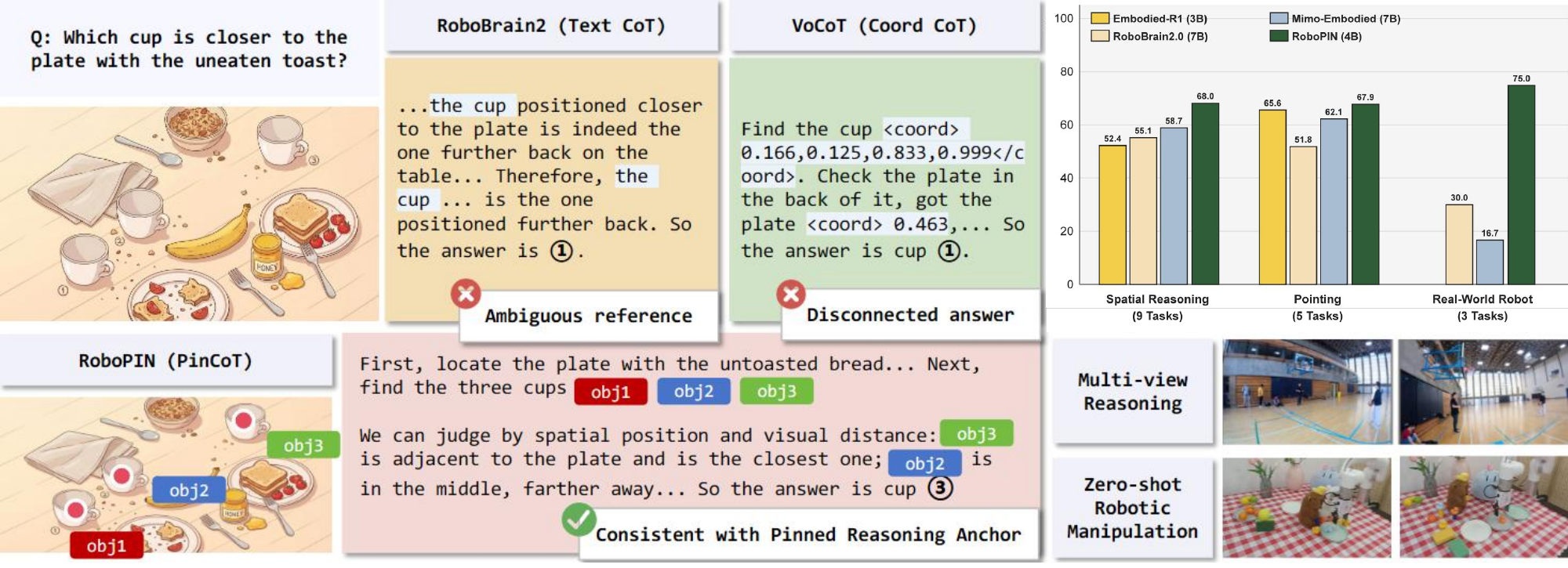}
    \caption{Comparison of three chain-of-thought paradigms for embodied reasoning. Text CoT relies on implicit textual descriptions with ambiguous entity references. Coord CoT embeds coordinates but lacks identity tracking, causing disconnection between reasoning and the final answer. \pincot{} (ours) pins each reasoning step to visual evidence via pinned reasoning anchors, ensuring identity-level traceability from the reasoning process to the final answer.}
    \label{fig:teaser}
\end{figure*}

\begin{itemize}[leftmargin=*,nosep]

\item We propose \textbf{\pincot{}}, introducing the \reasoninganchor{} concept that converts implicit visual attention during reasoning into explicit, trackable structured visual anchors, achieving cross-step and cross-view reasoning consistency.

\item We build a fully automated, high-quality data generation pipeline and construct the \dataset{} dataset, enabling large-scale generation of \pincot{}-format reasoning supervision.

\item We train \textbf{\method{}} with three-stage post-training and process-supervised rewards that jointly optimize reasoning process quality and answer correctness. Impressively, with only 4B parameters, \method{} achieves best average performance across 14 benchmarks spanning embodied spatial reasoning, pointing, and multi-view reasoning, outperforming 7B level open-source embodied models with a 12\% average improvement over the strongest 7B baseline Mimo-Embodied~\cite{hao2025mimo}.

\end{itemize}

\section{Related Work}

\subsection{Embodied Spatial Reasoning}

Spatial understanding is essential for embodied agents to bridge the ``seeing-to-doing'' gap.~\cite{yuan2025embodied}. To act reliably in physical environments, a model must not only recognize objects, but also infer their locations, spatial relations,
and task-relevant regions. Recent vision-language models have substantially improved this capability by incorporating explicit spatial coordinates, geometric reasoning, and scene-structure modeling into multimodal
representations~\cite{chen2024spatialvlm,cheng2024spatialrgpt}. In robotic manipulation settings, another important line of work predicts actionable points and visual traces that can directly support downstream control, such as grasping, placing, or interaction
planning~\cite{yuan2024robopoint,yuan2025seeing,qu2025spatialvla}. Together, these efforts have significantly strengthened embodied models in tasks such as spatial referring, target localization, and manipulation-oriented grounding.

\subsection{Embodied Multi-View Reasoning}

To alleviate the occlusion and partial visibility issues inherent to single-view observations, multi-view configurations have become a standard paradigm in embodied perception and manipulation~\cite{jangir2022look,jang2023murm}. Existing embodied foundation models~\cite{kim2024openvla,team2024octo} typically integrate these multi-view inputs through feature concatenation, cross-view attention mechanisms, or unified encoders to enhance overall scene coverage and spatial awareness.
However, the core challenge of multi-view reasoning extends beyond mere information fusion; it inherently demands robust \textit{cross-view correspondence}. A model must reliably determine whether distinct regions or entities across disparate views correspond to the same physical target. Recent multi-view benchmarks~\cite{feng2025seeing,wang2025towards,ray2024sat} reveal that current vision-language models remain highly fragile in this capability. When faced with drastic shifts in perspective, scale, and visibility, they often fail to consistently associate the same entity. The fundamental root cause is that most existing grounded reasoning methods predominantly produce localized, one-off spatial references, lacking a persistent identity-binding mechanism. Our work explicitly addresses this bottleneck by enforcing ID consistency constraints across multi-view observations, ensuring reliable and persistent entity alignment.


\subsection{Visually Grounded Chain-of-Thought}

Chain-of-Thought (CoT)~\cite{wei2022chain} significantly enhances the reasoning abilities of large language models. To extend this paradigm to multimodal domains, early works primarily focused on augmenting the intermediate reasoning process with visual evidence\cite{zhang2023multimodal,zhang2023multimodal,zheng2023ddcot}. While these methods successfully improve interpretability, their reasoning processes remain predominantly text-centric.

To better leverage visual information within reasoning, subsequent methods progressed by directly injecting spatial references into the reasoning chain\cite{chen2023shikra,you2023ferret,li2025vocot}. However, while these methods successfully make visual grounding explicit, they predominantly treat spatial references as localized, single-use annotations attached to individual reasoning steps. Consequently, entity tracking still relies on implicit natural language descriptions, leaving current models highly susceptible to cross-step reference drift. Our work addresses this limitation by introducing the \reasoninganchor{}. By representing grounded entities as persistent, identity-bound anchors, we make intermediate references trackable across steps.




\section{Method}
\label{sec:method}

This section presents the complete framework of \method{}. Section~\ref{sec:pincot} defines the \pincot{} structured reasoning representation, including the design of \reasoninganchor{}s, the identity tracking mechanism, and its natural extension to multi-view settings. Section~\ref{sec:data} describes a fully automated data construction pipeline for generating high-quality \pincot{}-format reasoning supervision at scale. Section~\ref{sec:training} introduces the three-stage progressive training strategy and the process-supervised reward design.

\subsection{Pinned Chain-of-Thought Representation}
\label{sec:pincot}

As discussed in the introduction, although existing paradigms have introduced visual coordinates into language models, these spatial references are typically isolated and lack explicit identity tracking. This often causes the reasoning trajectory to decouple from visual evidence, leading to reference drift and inconsistency across steps. To address the reasoning consistency problem in embodied reasoning, we propose Pinned Chain-of-Thought (\pincot{}), whose core concept is the \textit{\reasoninganchor{}}.

A \reasoninganchor{} is fundamentally different from conventional visual grounding. Traditional grounding operations such as object detection and referring expression comprehension, operate primarily at the perceptual level and are not explicitly maintained throughout multi-step reasoning. Once an object is localized, the grounding result is typically not reused as a persistent variable. In contrast, a \reasoninganchor{} functions as a reasoning-level variable that unifies identity and grounding. Upon introduction, it binds an entity to a unique identity along with its spatial location and view context, and persists as a variable throughout the reasoning process, forming a traceable and step-wise verifiable reasoning chain.

\pincot{} encodes entities as lightweight XML tags. Two tag types are defined: \texttt{obj} for physical objects and \texttt{space} for task-relevant spatial regions. Both share a unified parameterized schema:

\begin{tcolorbox}[colback=gray!10!white, colframe=gray!50!black, arc=1pt, boxrule=0.5pt, left=4pt, right=4pt, top=4pt, bottom=4pt, halign=center]
\small\ttfamily
<obj name="NAME" id="ID" img\_idx="VIEW" point="[x, y]">\\
<space name="NAME" id="ID" img\_idx="VIEW" point="[x, y]">
\end{tcolorbox}

Here, \texttt{name} denotes a semantic label, \texttt{id} is an explicit identity token, \texttt{img\_idx} indicates the source view, and \texttt{point} represents a 2D coordinate normalized to the range 0--1000. Formally, each \reasoninganchor{} can be represented as a structured tuple $a = (n, i, v, p)$. This representation design tightly binds semantic meaning, identity information, view provenance, and spatial localization into a unified representation.

We adopt point coordinates over bounding boxes for three reasons. First, many embodied task targets are inherently about ``which location'' rather than ``which region'' (e.g., pointing, affordance prediction, placement positions), and points naturally align with such operational targets. Second, points are more compact: they incur lower information overhead when used as anchors within a reasoning chain. Third, the combination of point and ID is better suited as a reasoning variable, analogous to a pointer in a program that refers to a precise location while carrying identity, whereas a bounding box resembles an output of a detection module. We validate this design choice through ablation experiments comparing point and bounding box representations in Section~\ref{sec:ablation}.

Building on these structured tags, \pincot{} enforces an identity-aware reasoning mechanism. When an entity first appears during reasoning, it must be introduced through a complete XML tag that establishes both its spatial localization and identity. Subsequent reasoning steps reference the same entity directly through its ID, without regenerating a full tag.

This mechanism naturally extends to multi-view settings. The same entity observed across viewpoints shares a unified identity token, varying only in \texttt{img\_idx} and \texttt{point}. By enforcing a shared identity, cross-view consistency becomes an explicit constraint within the reasoning process, thereby reducing identity drift and enabling consistent reasoning about the same entity from different observation angles.

\subsection{Automated Data Construction}
\label{sec:data}

Training models to follow \pincot{} requires aligning visual grounding and identity with the reasoning process. However, existing multimodal datasets rarely provide such unified annotations, and directly converting raw data into \pincot{} format using large models tends to bypass explicit intermediate grounding steps, resulting in anchors that are not reliably grounded in visual evidence. We therefore build a fully automated three-stage pipeline organized around explicit structured intermediate representations, incorporating dedicated quality assurance throughout the process. Using this pipeline, we construct the \dataset{} reasoning dataset.

To identify task-relevant entities, we first use \teacher{}~\cite{bai2025qwen3} to analyze the current question and extract semantic descriptions of candidate entities. Since embodied task types vary significantly, a single prompting strategy cannot adequately cover all cases. We therefore partition the data into several task categories (e.g., spatial relation reasoning, pointing, affordance prediction) and design task-specific prompts for semantic parsing. This targeted strategy produces stable semantic specifications that accurately capture target entities.

Given the parsed semantic metadata, we employ a Florence-2~\cite{xiao2024florence} + SAM~2.1~\cite{ravi2024sam} pipeline to obtain precise point coordinates and bounding boxes for the relevant entities. This stage retains the predicted bounding boxes, refined segmentation masks, and derived point anchors as structured grounding evidence for subsequent use. However, this pipeline is not equally reliable across all target types: for small symbolic targets such as ``X'' marks in images, Florence-2 + SAM~2.1 often fails to localize accurately. To address this, we introduce a reverse prediction verification step, where Florence is asked to identify what the predicted point corresponds to. Only predictions whose reverse interpretation is consistent with the original target description are retained, substantially improving the quality and reliability of the grounding annotations.

We then feed the structured grounding evidence as ground-truth information into \teacher{} to generate the corresponding \pincot{} reasoning chains. To improve generation quality, we impose task-specific constraints for different task categories, reducing arbitrary or unsupported reasoning during generation. After obtaining \pincot{} outputs, we further apply rule-based filtering to remove low-quality reasoning chains and inconsistent results, such as incorrect ID references or mismatched anchor coordinates, ensuring the overall quality of the constructed data.

Through this three-stage pipeline, multimodal datasets are progressively transformed into structured reasoning data, in which visual evidence, entity identity, provenance, and answer supervision remain explicitly aligned throughout the construction process. The entire pipeline executes fully automatically without manual annotation and scales to large volumes of data, yielding the \dataset{} reasoning dataset.
Further details regarding the pipeline can be found in Appendix A.

\subsection{Progressive Training Pipeline}
\label{sec:training}

Building on the framework and data, we start from \student{}~\cite{bai2025qwen3} and progressively inject the capabilities through three-stage post-training: Stage~1 performs embodied domain adaptation via SFT, Stage~2 learns \pincot{} structured reasoning via CoT-SFT, and Stage~3 achieves process-supervised alignment via RFT.


We train \method{} on a diverse corpus to improve spatial understanding, multi-view reasoning, pointing, and long-horizon embodied reasoning. Different data subsets are assigned to different training stages according to their supervision signals and targeted capabilities. For Stage~2, we additionally apply the data construction pipeline described in Section~\ref{sec:data} to generate structured \pincot{} samples for learning \reasoninganchor{} introduction and the identity-aware reasoning mechanism.

\textbf{Geometric and general spatial reasoning data.}
This subset is primarily derived from \textbf{Euclid}~\cite{lian2025euclid} and \textbf{Video-R1}~\cite{feng2025video}. Euclid provides structured problems involving spatial relations and geometric properties, helping the model build precise spatial reasoning capabilities. Video-R1 provides general visual understanding samples from dynamic and diverse scenes, preserving the model's broad perceptual abilities during training.

\textbf{Embodied spatial reasoning and pointing data.}
This subset is primarily derived from \textbf{EmbSpatial}~\cite{du2024embspatial}, \textbf{EO-Data}~\cite{qu2025embodiedonevision}, \textbf{RoboVQA}~\cite{sermanet2024robovqa}, \textbf{EgoPlan}~\cite{chen2026egoplan}, and \textbf{Embodied-Point}~\cite{yuan2025embodied}. EmbSpatial and EO-Data focus primarily on embodied spatial perception, driving the model to develop robust environment-level reasoning capabilities. RoboVQA and EgoPlan textitasize task-oriented and egocentric embodied scenarios, introducing action-related reasoning and sequential decision-making processes. Furthermore, Embodied-Point covers diverse multi-task pointing scenarios, helping the model learn to generate precise point coordinates in embodied environments.

\textbf{Multi-view understanding data.}
This subset is primarily derived from \textbf{CrossPoint}~\cite{wang2025towards} and \textbf{SAT}~\cite{ray2024sat}, and augmented with additional multi-view data constructed from \textbf{InternData-M1}~\cite{contributors2025internroboticsrepo}. CrossPoint and SAT provide observations of the same scene under diverse camera viewpoints, enabling the model to learn consistent object representations and spatial relationships across views. InternData-M1 provides a diverse array of scenes, serving as a crucial foundation for multi-view reasoning in embodied tasks.

Detailed statistics of the training corpus, including the size of each data subset and the mixture ratios used across progressive training stages, are provided in Appendix B.


The first stage focuses on embodied domain adaptation. We fine-tune the model on embodied data covering object understanding, egocentric activity recognition, pointing and planning, enabling it to build foundational understanding of objects, affordances, spatial relations, and action events. A small amount of general-domain data is mixed in to mitigate catastrophic forgetting. The second stage injects \pincot{} structured reasoning ability. We leverage the pipeline from Section~\ref{sec:data} to generate structured, multi-step \pincot{} reasoning traces for supervised learning. In this stage, the model learns two key capabilities: introducing entities through complete XML tags to establish their \reasoninganchor{}s, and maintaining consistent identity references across both multi-step reasoning chains and multiple viewpoints.


The third stage focuses on process-supervised alignment. After the first two stages, the model can generate structured reasoning chains, but may still prefer fluent yet poorly anchored outputs during inference. To counteract this tendency, we apply GRPO~\cite{shao2024deepseekmath} with a composite reward function that simultaneously supervises both the reasoning process and the final answer.

For a generated response $y$, let $\hat{a}$ denote the parsed final answer, $\mathcal{M}(y)$ the set of entity mentions appearing in the \texttt{<think>} block, and $\mathcal{M}_{\text{valid}}(y) \subseteq \mathcal{M}(y)$ the subset whose first occurrence is introduced by a valid format and whose subsequent mentions strictly maintain cross-step identity consistency. We further denote by $\mathcal{P}_{\text{xml}}(y)$ the set of anchor points extracted from the reasoning trace. For certain specific tasks, each predicted point $p_j$ is paired with a target grounding region $b_j^\star$.

We design a composite reward that evaluates four complementary dimensions of output quality:
\begin{equation}
R = \lambda_f R_{\text{format}} + \lambda_c R_{\text{consistency}} + \lambda_p R_{\text{pin}} + \lambda_a R_{\text{accuracy}},
\end{equation}
where $\lambda_\cdot$ are task-specific weights. When a reward term is not applicable to a particular task, its weight is set to zero.

\textbf{Format reward} $R_{\text{format}}$ evaluates whether the output structure is well-formed. The reasoning process must be enclosed within \texttt{<think>...</think>} and the answer within the format of \texttt{<answer>}...\texttt{</answer>}. The \texttt{<think>} block must also contain valid \reasoninganchor{} tags. The highest reward is assigned when both formatting and \reasoninganchor{} constraints are satisfied, a lower reward when only the outer format is correct, and zero otherwise.

\textbf{Consistency reward} $R_{\text{consistency}}$ evaluates the coherence of entity references throughout the reasoning process, measuring the proportion of entity mentions that follow the identity-aware reasoning mechanism:
\begin{equation}
R_{\text{consistency}} =
\frac{|\mathcal{M}_{\text{valid}}(y)|}{\max(|\mathcal{M}(y)|, 1)}.
\end{equation}
This reward encourages the model to maintain stable entity references across reasoning chains and multiple views, reducing identity drift and reference ambiguity.

\textbf{Pin reward} $R_{\text{pin}}$ evaluates the localization accuracy of \reasoninganchor{}s in the reasoning process. This term is applied only to datasets with explicit target region supervision, avoiding over-constraining the model's reasoning on other tasks:
\begin{equation}
R_{\text{pin}} =
\frac{\left| \left\{ p_j \in \mathcal{P}_{\text{xml}}(y) \mid p_j \in b_j^\star \right\} \right|}{\max(|\mathcal{P}_{\text{xml}}(y)|, 1)}.
\end{equation}
By directly supervising the localization accuracy of intermediate anchors, the pin reward ensures that \reasoninganchor{}s during reasoning are genuinely ``pinned'' to the correct locations.

\textbf{Accuracy reward} $R_{\text{accuracy}}$ evaluates the correctness of the final answer. Since answer formats vary across tasks, this reward is instantiated in a task-dependent manner: for multiple-choice tasks, it is a binary indicator of exact match; for pointing tasks, it measures the proportion of predicted points falling within the target bounding boxes; for open-ended tasks, it uses normalized sentence-level BLEU~\cite{papineni2002bleu} as a soft text-matching reward.

Overall, the reward design introduces process supervision over \textit{how the model reasons}. $R_{\text{format}}$ ensures well-formed outputs, $R_{\text{consistency}}$ enforces stable identity-aware reasoning mechanism, $R_{\text{pin}}$ directly supervises \reasoninganchor{} localization quality, and $R_{\text{accuracy}}$ ensures final-answer correctness. Together, these four terms enforce reasoning quality from both the process and the final answer, ultimately driving the model to generate high-quality \pincot{} reasoning chains. Further details regarding the training configurations can be found in Appendix C.   


\section{Experiments}

We evaluate \method{} on embodied spatial reasoning (Sec.~\ref{sec:embodied_reasoning}), point-level grounding (Sec.~\ref{sec:pointing}), and real-world robot execution (Sec.~\ref{sec:real_world}). We then ablate the contributions of \pincot{} and the three-stage training recipe (Sec.~\ref{sec:ablation}), and analyze how process supervision improves reasoning quality and consistency (Sec.~\ref{sec:reasoning_analysis}). Finally, we verify that our embodied alignment preserves general vision-language capabilities without catastrophic forgetting, with detailed results provided in Appendix D.

\subsection{Embodied Cognition and Spatial Reasoning}
\label{sec:embodied_reasoning}
\textbf{Setup.} We evaluate the embodied cognition and spatial reasoning capabilities of our model across 9 benchmarks encompassing 13 distinct capability dimensions. These include ERQA~\cite{team2025gemini}, CV-Bench~\cite{tong2024cambrian}, EmbSpatial~\cite{du2024embspatial}, SAT~\cite{ray2024sat}, RoboSpatial~\cite{song2025robospatial}, Robo-VQA~\cite{sermanet2024robovqa}, CrossPoint~\cite{wang2025towards}, MVRoboBench~\cite{feng2025seeing}, and BLINK~\cite{fu2024blink}. These benchmarks comprehensively cover various aspects of spatial understanding, object localization, and physical reasoning. To rigorously assess performance, we compare \method{} against two leading closed-source models: GPT-5~\cite{singh2025openai} and Gemini-2.5-Pro~\cite{comanici2025gemini}, alongside seven competitive open-source vision-language models equipped with advanced spatial reasoning and grounding capabilities. These include generalist multi-modal models (Qwen3-VL~\cite{bai2025qwen3} and InternVL3.5~\cite{wang2025internvl3}), as well as recent domain-specific models designed for embodied reasoning (RoboBrain2.0~\cite{team2025robobrain}, VeBrain~\cite{luo2025visual}, Mimo-Embodied~\cite{hao2025mimo}, Pelican-VL~\cite{zhang2025pelican}, and Embodied-R1~\cite{yuan2025embodied}).

\textbf{Results.} As detailed in Table~\ref{tab:main_reasoning_results}, \method{} demonstrates exceptional general spatial reasoning capabilities. It achieves the best open-source results on EmbSpatial, SAT, and RoboSpatial, highlighting the strong advantage of \pincot{} in tasks that require comparing relative object positions and tracking spatial changes. Across the other embodied cognition benchmarks, our model consistently performs within the first tier. Furthermore, on multi-view specific evaluations, \method{} ranks first on CrossPoint, as well as the Multi-View and Visual Correspondence subsets of BLINK. This indicates that the universal identity design across views effectively facilitates stable object correspondence, enhancing the model's multi-view reasoning proficiency.

\begin{table*}[t]
  \centering
  \scriptsize
  \setlength{\tabcolsep}{2.8pt}
  \caption{Performance on spatial reasoning benchmarks. \textbf{Bold} and \underline{underlined} denote the best and second open-source results.}
  \label{tab:main_reasoning_results}
  \resizebox{\textwidth}{!}{%
  \begin{tabular}{lcccccccc|cc|cccc}
  \toprule
  \multirow{2}{*}{Model} & \multirow{2}{*}{Params} & \multirow{2}{*}{ERQA} & \multirow{2}{*}{\shortstack{CV-Bench}} & \multirow{2}{*}{\shortstack{EmbSpatial}} & \multirow{2}{*}{SAT} & \multirow{2}{*}{\shortstack{RoboSpatial}} & \multirow{2}{*}{\shortstack{RoboVQA}} &
  \multirow{2}{*}{CrossPoint} & \multicolumn{2}{c|}{\textbf{MV-RoboBench}} & \multicolumn{4}{c}{\textbf{BLINK}} \\
  & & & & & & & & & CrossV. Match & Traj. Sel & Multi. View & \shortstack{Vis. Corr.} & \shortstack{Spat. Rel.} & \shortstack{Rel. Depth} \\
  \midrule
  \rowcolor{gray!10} \multicolumn{15}{l}{\textit{Closed-source models}} \\
  GPT-5 & - & 54.5 & 84.5 & 78.3 & 83.3 & 54.1 & 33.1 & 33.0 & 29.0 & 54.5 & 45.8 & 77.3 & 90.2 & 82.3 \\
  Gemini-2.5-Pro & - & 55.7 & 84.6 & 78.7 & 76.7 & 59.9 & 33.9 & 37.1 & 39.5 & 65.5 & 46.6 & 78.5 & 85.3 & 86.3 \\
  \rowcolor{gray!10} \multicolumn{15}{l}{\textit{Open-source generalist models}} \\
  Qwen3-VL & 4B & 41.8 & 85.0 & \underline{77.1} & 70.7 & 59.4 & 41.1 & 29.3 & \underline{24.0} & 31.0 & \underline{51.1} & \underline{88.4} & \underline{84.6} & 85.5 \\
  InternVL3.5 & 8B & 41.0 & 81.5 & 70.3 & 55.3 & 51.1 & 28.6 & 19.2 & 21.5 & \underline{35.0} & 47.4 & 72.7 & 81.8 & 78.2 \\
  \rowcolor{gray!10} \multicolumn{15}{l}{\textit{Open-source embodied models}} \\
  RoboBrain2.0 & 7B & 38.5 & \underline{85.8} & 76.3 & 75.3 & 54.2 & 57.5 & 26.0 & 21.0 & 26.0 & 48.1 & 46.5 & 80.4 & 80.7 \\
  VeBrain & 7B & 37.3 & 79.7 & 70.5 & 58.0 & 42.5 & 42.4 & 20.2 & 18.0 & 30.0 & 46.6 & 48.8 & 83.2 & 67.7 \\
  Mimo-Embodied & 7B & \textbf{46.8} & \textbf{88.8} & 76.2 & 54.6 & \underline{61.8} & \textbf{62.0} & \underline{36.8} & 21.1 & 34.0 & 26.3 & 83.2 & 77.6 & \textbf{94.4} \\
  Pelican-VL & 7B & 39.8 & 78.9 & 73.2 & 54.6 & 57.5 & \underline{58.5} & 23.8 & 23.0 & 32.5 & 47.3 & 63.3 & \underline{84.6} & 75.0 \\
  Embodied-R1 & 3B & 35.2 & 82.7 & 67.4 & \underline{76.3} & 47.4 & 51.8 & 25.2 & 21.5 & 26.5 & 40.6 & 50.0 & 76.9 & 79.8 \\
  \midrule
  \rowcolor{blue!10} \method{} & 4B & \underline{44.0} & 85.5 & \textbf{83.1} & \textbf{78.0} & \textbf{66.6} & 56.1 & \textbf{72.2} & \textbf{26.0} & \textbf{35.5} & \textbf{72.9} & \textbf{89.5} & \textbf{87.4} & \underline{87.1} \\
  \bottomrule
  \end{tabular}%
  }
\end{table*}

\subsection{Embodied Pointing and Location}
\label{sec:pointing}
\textbf{Setup.} We next examine point-level grounding ability by evaluating on five distinct benchmarks: VABench-P~\cite{yuan2025seeing}, Where2Place~\cite{yuan2024robopoint}, RefSpatial~\cite{zhou2025roborefer}, RoboRefit~\cite{lu2023vl}, and RoboAfford~\cite{tang2025roboafford}. 

\textbf{Results.} As shown in Table~\ref{tab:main_pointing_results}, \method{} achieves the best open-source performance on RefSpatial, RoboRefit, and RoboAfford. We note that Embodied-R1, as a model specifically designed and trained for pointing tasks, shows strong performance on VABench-P and Where2Place. Despite being a general-purpose embodied reasoning model, \method{} remains competitive on two benchmarks while significantly outperforming Embodied-R1 on broader spatial reasoning tasks, demonstrating a more balanced capability profile. 

\begin{table}[t]
  \centering
  \scriptsize
  \setlength{\tabcolsep}{6pt}
  \caption{Performance on pointing benchmarks. \textbf{Bold} and \underline{underlined} values denote the best and second-best open-source results.}
  \label{tab:main_pointing_results}
  \begin{adjustbox}{max width=\columnwidth}
  \begin{tabular}{lcccccc}
    \toprule
    Model & Params & \shortstack{VABench-P} & \shortstack{Where2Place} & \shortstack{Ref-Spatial} & \shortstack{Robo-Refit} & \shortstack{Robo-Afford} \\
    \midrule
    \rowcolor{gray!10} \multicolumn{7}{l}{\textit{Closed-source models}} \\
    GPT-5 & - & 31.1 & 39.7 & 21.6 & 44.7 & 30.0 \\
    Gemini-2.5-Pro & - & 21.7 & 49.6 & 36.5 & 38.4 & 23.4 \\
    \rowcolor{gray!10} \multicolumn{7}{l}{\textit{Open-source generalist models}} \\
    Qwen3-VL & 4B & 30.6 & \underline{67.7} & 43.0 & \underline{86.1} & \underline{70.1} \\
    InternVL3.5 & 8B & 23.0 & 34.8 & 16.8 & 30.8 & 31.5 \\
    \rowcolor{gray!10} \multicolumn{7}{l}{\textit{Open-source embodied models}} \\
    RoboBrain2.0 & 7B & 41.0 & 63.6 & 32.5 & 70.4 & 51.5 \\
    VeBrain & 7B & 1.7 & 12.3 & 0.3 & 32.2 & 2.1 \\
    Mimo-Embodied & 7B & 46.9 & 63.6 & \underline{48.0} & 82.3 & 69.8 \\
    Pelican-VL & 7B & 14.5 & 57.8 & 37.5 & 74.9 & 63.4 \\
    Embodied-R1 & 3B & \textbf{66.0} & \textbf{69.5} & 39.7 & 85.6 & 67.2 \\
    \midrule
    \rowcolor{blue!10} \method{} & 4B & \underline{65.0} & 62.0 & \textbf{50.5} & \textbf{89.0} & \textbf{72.8} \\
    \bottomrule
  \end{tabular}%
  \end{adjustbox}
\end{table}

\begin{table*}[t]
  \begin{minipage}[t]{0.49\textwidth}
    \vspace{0pt}
    \centering
    \scriptsize
    \setlength{\tabcolsep}{5pt}
    \captionof{table}{Reasoning process analysis on EmbSpatial.
     A-Cov: Anchor Coverage; AC: Anchor Correctness; CS-IDCov: Cross-Step identity Coverage.}
    \label{tab:reasoning_process_combined}
    \begin{adjustbox}{max width=\linewidth,center}
      \begin{tabular}{lccc}
        \toprule
        Model & A-Cov (\%) & AC (\%) & CS-IDCov (\%) \\
        \midrule
        \method{} &  \textbf{100.0} & \textbf{89.1} & \textbf{99.9} \\
        \hspace{0.8em}w/o RFT  & 96.3 & 87.0 & 18.8 \\
        \bottomrule
      \end{tabular}
    \end{adjustbox}
  \end{minipage}
  \hfill
  \begin{minipage}[t]{0.48\textwidth}
    \vspace{0pt}
    \centering
    \scriptsize
    \captionof{table}{Error propagation on EmbSpatial.}
    \label{tab:anchor_answer_risk}
    \begin{adjustbox}{max width=\linewidth,center}
      \begin{tabular}{lccc}
        \toprule
        & \multicolumn{2}{c}{Answer Error Rate (\%)} & \\
        \cmidrule(lr){2-3}
        Model & w/ Faulty Anchor & w/ Correct Anchor & Risk Ratio \\
        \midrule
        \method{} & 29.2 & 13.2 & 2.2 \\
        \bottomrule
      \end{tabular}
    \end{adjustbox}
  \end{minipage}
\end{table*}

\begin{table}[t]
\caption{Cross-view reasoning consistency on CrossPoint. Acc: Overall Accuracy; ID-Trans: Shared-ID Transfer Rate; Trans-Acc: Accuracy given successful ID transfer.}
\label{tab:crosspoint_consistency}
\centering
\small
\setlength{\tabcolsep}{8pt}
\begin{tabular}{lccc}
  \toprule
  Model & Acc (\%) & ID-Trans (\%) & Trans-Acc (\%) \\
  \midrule
   \method{} & \textbf{71.1} & \textbf{86.3} & \textbf{72.1} \\
  \hspace{0.8em}w/o RFT & 62.5 & 60.6 & 61.1 \\
  \bottomrule
\end{tabular}
\end{table}

\subsection{Real-World Robot Evaluation}
\label{sec:real_world}
We conduct zero-shot real-world evaluations using an xArm robot. The system is equipped with two RGB cameras: a static camera observing the tabletop scene and a wrist-mounted camera. Given a language instruction, the model predicts target points via visual grounding, which are then executed using the point-based motion planning pipeline from Embodied-R1~\cite{yuan2025embodied}.

\textbf{Tasks.} We design three types of complex manipulation tasks to evaluate spatial reasoning, multi-view understanding, and long-horizon planning:

\textbf{(1) Spatial reasoning task}: \textit{“Pick up the duck in front of two toys and place it to the middle plate among plates.”}

\textbf{(2) Multi-view reasoning task}: \textit{“Pick up the duck behind two toys and place it to the plate in front of the two toys.”}

\textbf{(3) Long-horizon task}: \textit{“Pick up yellow objects located outside the plates and place into the yellow plate.”}

To rigorously ensure robustness, we extensively randomize object instances, distractor placements, and overall scene layouts for each task. We conduct 20 distinct trials per setting under these varied conditions to compute the success rate.

\begin{figure*}[t]
  \centering
  \includegraphics[width=\textwidth]{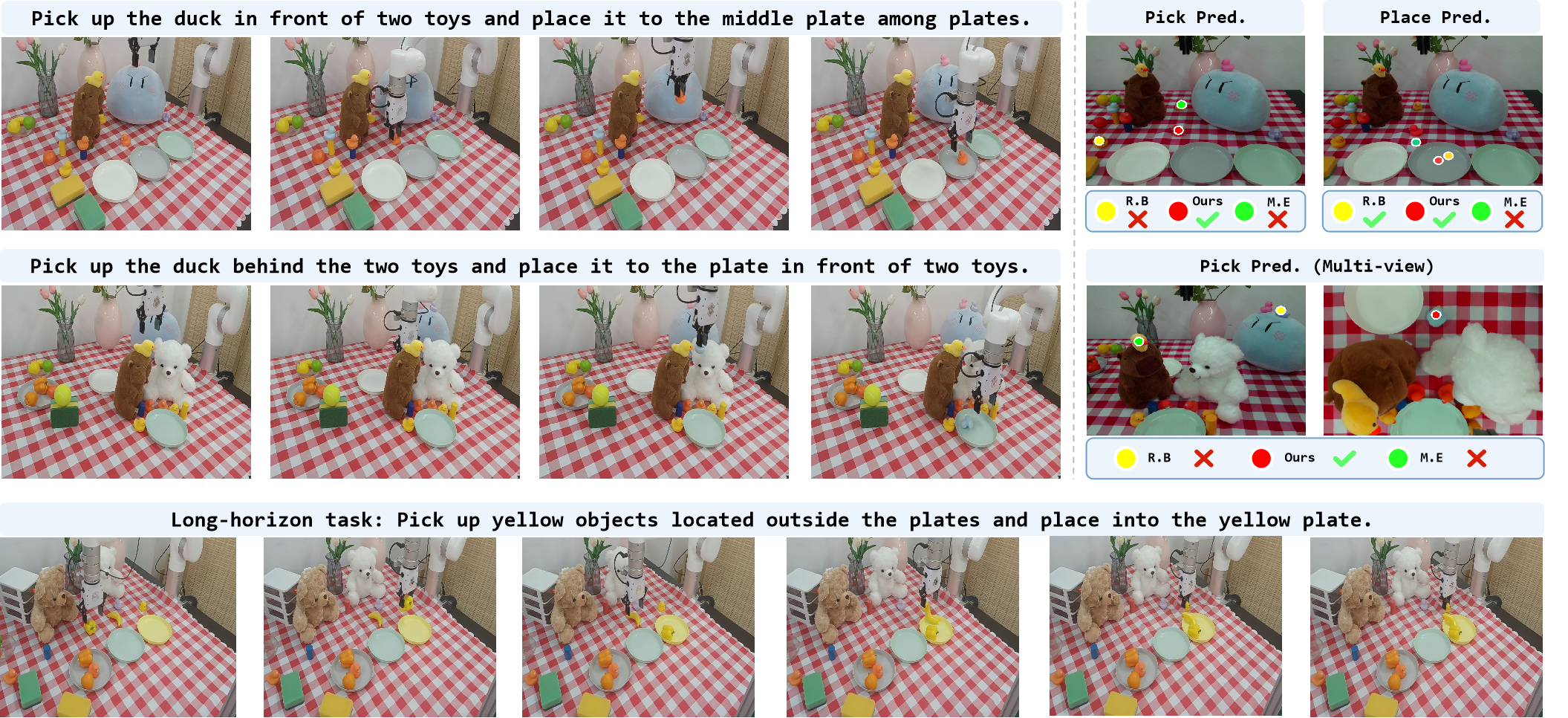}
  \caption{Real-world robot visualization. Left: robot execution. Right: predicted anchor points during manipulation. R.B and M.E denote RoboBrain2.0 and Mimo-Embodied.}
  \label{fig:realworld_robot}

  \vspace{0.5em}

  \captionof{table}{Real-world robot evaluation. Each setting is evaluated over 20 independent trials.}
  \label{tab:realworld_results}
  \centering
  \small
  \begin{tabular}{lccc}
    \toprule
    \textbf{Task} 
    & \shortstack{RoboBrain2.0} & \shortstack{Mimo-Embodied} & \shortstack{\method{}} \\
    \midrule
      Spatial         & 12/20 & 7/20  & \textbf{20/20} \\
      Multi-view      & 1/20  & 0/20  & \textbf{14/20} \\
      Long-horizon    & 5/20  & 3/20  & \textbf{11/20} \\
    \bottomrule
  \end{tabular}
\end{figure*}

\textbf{Results.} Table~\ref{tab:realworld_results} reports the success rates of different methods, including RoboBrain2-7B and Mimo-Embodied-7B. In multi-view tasks, baseline models tend to rely solely on the static camera and often select incorrect targets, failing to utilize the wrist-view information. In contrast, our model effectively integrates multi-view observations and demonstrating enhanced reliability in identifying the target object compared to the baselines.

In spatial reasoning tasks, our model also achieves consistently higher accuracy, reflecting a stronger capacity for interpreting spatial relations and resolving ambiguous object configurations. For the long-horizon task, RoboPIN predicts keypoints step-by-step and outperforms other models at each step, leading to a substantially higher overall success rate. Figure~\ref{fig:realworld_robot} illustrates representative execution sequences alongside the predicted anchor points for multi-view and spatial tasks, visually confirming the model's ability to ground its reasoning in precise spatial locations.

\subsection{Ablation Study}
\label{sec:ablation}

In this section, we conduct ablation studies on three key components: the reasoning format, the reward design, and the three-stage training pipeline.
For the format and reward ablations, we use identical training hyperparameters on fixed training subsets to isolate the effects of reasoning format and reward design.

\textbf{Performance Comparison of Reasoning Formats.} We compare four spatial representations: standard text-only CoT (\textit{Text CoT}), coordinate-augmented CoT (\textit{Coord CoT}), coordinate CoT with explicit IDs (\textit{Coord CoT w/ ID}), and our proposed \pincot{} which utilizes point-based anchors with IDs. As shown in Table~\ref{tab:format_rft_ablation}, introducing explicit reasoning generally improves upon the non-reasoning baseline, but the spatial representation plays a critical role. Across the two benchmarks, \pincot{} generally outperforms other formats, and its advantage is significantly amplified with the introduction of RFT. Notably, under standard SFT, \pincot{} underperforms \textit{Coord CoT} on EmbSpatial. However, once RFT is applied, \pincot{} successfully surpasses \textit{Coord CoT}. This reversal suggests that \pincot{} provides a more stable and effective reasoning interface, but requires RFT to fully realize its potential. Furthermore, comparing \pincot{} with \textit{Coord CoT w/ ID}, which shares the same ID tracking mechanism but uses box coordinates instead of points, \pincot{} consistently outperforms it under both SFT and RFT settings. This confirms that point-based anchors serve as a more effective spatial primitive for reasoning, likely due to their lightweight and action-oriented nature that aligns better with embodied tasks.

\begin{figure*}[!t]
  \centering
  \includegraphics[width=\textwidth]{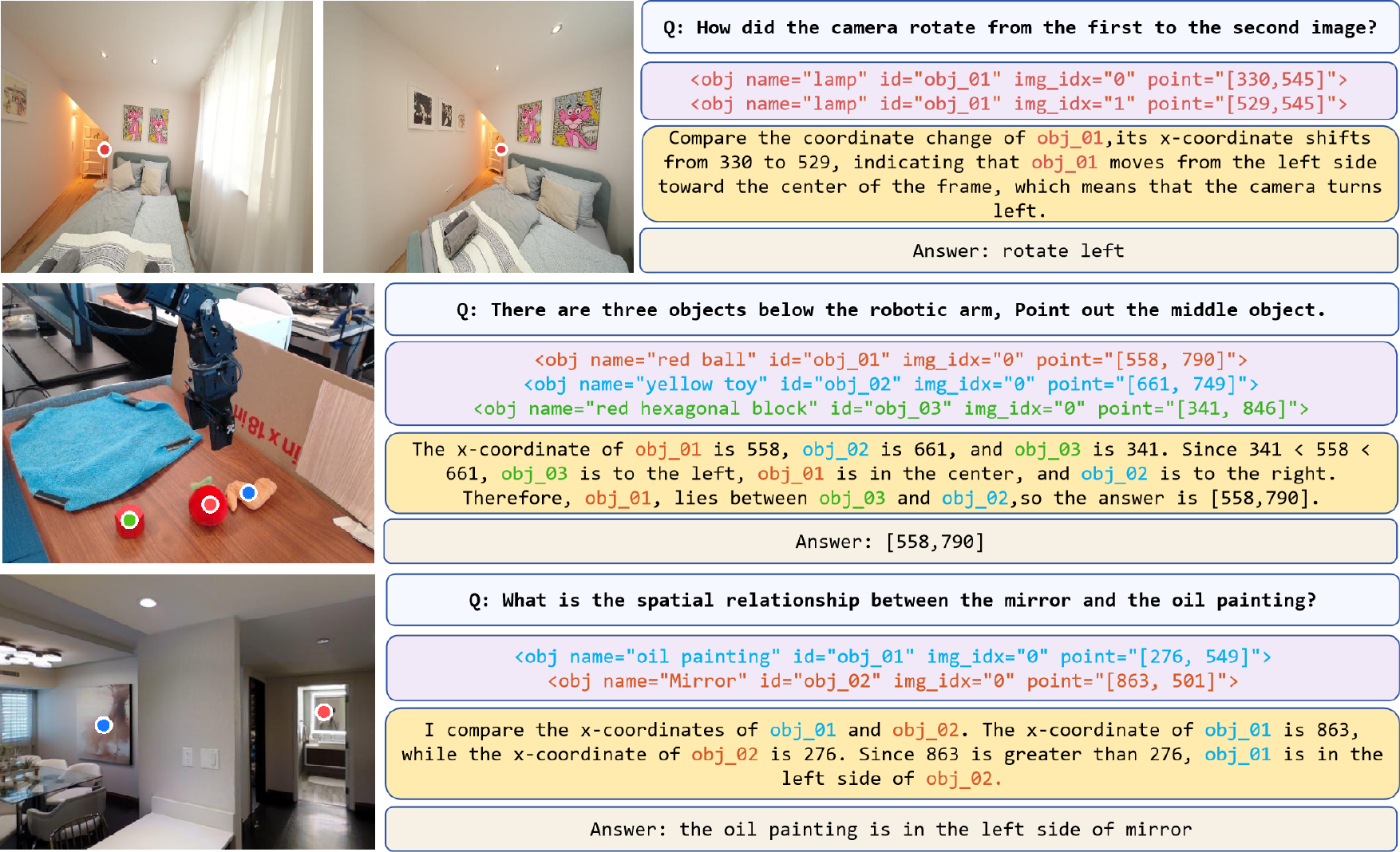}
  \caption{Qualitative analysis of grounded reasoning.}
  \label{fig:qualitative_reasoning}
\end{figure*}


\textbf{Ablation of Reward Components.}
We further ablate the two core reward components in RFT: the Pin reward for anchor localization and the Consistency reward for entity references. As shown in Table~\ref{tab:reward_ablation}, removing either component leads to a clear performance drop, confirming that both spatially accurate anchors and identity-consistent reasoning are necessary for strong embodied reasoning performance. Notably, removing the Pin reward causes a larger degradation ($-$2.7) than removing the Consistency reward ($-$2.0), suggesting that anchor localization quality is the more fundamental bottleneck for embodied reasoning.


\textbf{Performance Comparison of the Three-Stage Training.}
We further analyze how model performance progresses across different training stages based on average scores over embodied reasoning and pointing tasks. As shown in Table~\ref{tab:three_stage_ablation}, embodied reasoning improves progressively as additional stages are introduced: CoT-SFT yields a +5.0 gain over SFT alone, and RFT provides a further +5.7 improvement. This suggests that our three-stage recipe primarily strengthens high-level embodied spatial reasoning, while effectively preserving the fundamental point-level localization capabilities.


\begin{table*}[t]
  \centering
  \scriptsize
  \setlength{\tabcolsep}{4pt}
  \caption{Ablation of reasoning format and RFT. The first row indicates the Qwen3-VL-4B-Instruct baseline. All compared variants use the same training recipe and data subsets.}
  \label{tab:format_rft_ablation}
  \begin{adjustbox}{max width=\textwidth,center}
    \begin{tabular}{lcc|cc}
      \toprule
      \multicolumn{3}{c|}{\textbf{Reasoning Design}} & \multicolumn{2}{c}{\textbf{Performance}} \\
      \cmidrule(lr){1-3} \cmidrule(lr){4-5}
      Format & SFT & RFT & EmbSpatial & SAT \\
      \midrule
      Baseline & $\times$ & $\times$ & 77.1 & 70.7 \\
      Text CoT & $\checkmark$ & $\times$ & 78.9 & 74.0 \\
      Text CoT & $\checkmark$ & $\checkmark$ & 79.3 & 76.3 \\
      Coord CoT & $\checkmark$ & $\times$ & 81.7 & 67.3 \\
      Coord CoT & $\checkmark$ & $\checkmark$ & 83.1 & 74.0 \\
      Coord CoT w/ ID & $\checkmark$ & $\times$ & 78.6 & 70.7 \\
      Coord CoT w/ ID & $\checkmark$ & $\checkmark$ & 81.5 & 76.0 \\
      \pincot{} & $\checkmark$ & $\times$ & 80.3 & 74.0 \\
      \pincot{} & $\checkmark$ & $\checkmark$ & \textbf{84.7} & \textbf{78.7} \\
      \bottomrule
    \end{tabular}
  \end{adjustbox}
\end{table*}

\begin{table*}[t]
  \begin{minipage}[t]{0.38\textwidth}
    \vspace{0pt}
    \centering
    \scriptsize
    \captionof{table}{Ablation of reward components on EmbSpatial.}
    \label{tab:reward_ablation}
    \begin{adjustbox}{max width=\linewidth,center}
      \begin{tabular}{lc}
        \toprule
        Setting & EmbSpatial \\
        \midrule
        \method{} & \textbf{84.7} \\
        \hspace{0.8em}w/o Pin reward & 82.0 \\
        \hspace{0.8em}w/o Consistency reward & 82.7 \\
        \bottomrule
      \end{tabular}
    \end{adjustbox}
  \end{minipage}
  \hfill
  \begin{minipage}[t]{0.59\textwidth}
    \vspace{0pt}
    \centering
    \scriptsize
    \captionof{table}{Ablation of the three-stage training recipe.}
    \label{tab:three_stage_ablation}
    \begin{adjustbox}{max width=\linewidth,center}
      \begin{tabular}{ccc|cc}
        \toprule
        \multicolumn{3}{c|}{\textbf{Training Stages}} & \multicolumn{2}{c}{\textbf{Avg-Performance}} \\
        \cmidrule(lr){1-3} \cmidrule(lr){4-5}
        SFT & CoT-SFT & RFT & Spatial & Point \\
        \midrule
        $\checkmark$ & $\times$ & $\times$ & 57.0 & 66.0 \\
        $\checkmark$ & $\checkmark$ & $\times$ & 62.0 & 63.6 \\
        $\checkmark$ & $\checkmark$ & $\checkmark$ & \textbf{67.7} & \textbf{67.9} \\
        \bottomrule
      \end{tabular}
    \end{adjustbox}
  \end{minipage}
\end{table*}

\subsection{Reasoning Process Analysis}
\label{sec:reasoning_analysis}

To understand why \method{} achieves superior performance, we analyze the intermediate reasoning process on EmbSpatial (which provides ground-truth bounding boxes) and CrossPoint (for cross-view evaluation).

\textbf{Anchor Quality and Identity Consistency.} As shown in Table~\ref{tab:reasoning_process_combined}, \method{} achieves full anchor coverage (100.0\%) and high anchor correctness (89.1\%), both of which degrade without RFT. However, the key difference lies in identity tracking: without RFT, the model rarely sustains ID-based reasoning across steps (18.8\%) despite reasonably accurate anchors, whereas with RFT it achieves 99.9\% cross-step coverage. This shows that process supervision enforces a stable identity-aware reasoning mechanism beyond merely improving anchor quality. We further visualize the inference process on representative examples in Figure~\ref{fig:qualitative_reasoning}, confirming that such identity-aware tracking eliminates ambiguous textual references and ensures strong cross-step and cross-view consistency.


\textbf{Error Propagation from Reasoning to Answer.} As shown in Table~\ref{tab:anchor_answer_risk}, samples with faulty anchors exhibit roughly $2.2\times$ higher answer-error rates than those with correct anchors, confirming that reliable spatial grounding is a critical prerequisite for robust reasoning.


\textbf{Cross-view Consistency.} We evaluate the Pointing and Judgement subsets of CrossPoint, measuring overall accuracy (\textit{Acc}), shared-ID transfer rate across views (\textit{ID-Trans}), and conditional accuracy given successful transfer (\textit{Trans-Acc}). As shown in Table~\ref{tab:crosspoint_consistency}, \method{} consistently maintains shared-identity transfer across views (86.3\%), and when this transfer succeeds, predictions become highly reliable (72.1\%). Removing RFT substantially weakens this behavior (ID-Trans drops to 60.6\%), accompanied by a clear performance decline. These findings confirm that process supervision is the critical driver enforcing explicit cross-view correspondence.

\section{Conclusion}

We present Pinned Chain-of-Thought (PinCoT), a structured reasoning paradigm that pins every reasoning step onto visual evidence by unifying semantic name and spatial localization into a structured reasoning variable, enabling persistent entity tracking across reasoning steps and viewpoints. We build a fully automated data generation pipeline to construct the PIN-170K reasoning dataset. Then we train RoboPIN through three-stage progressive post-training, where the RFT stage employs a composite reward function that simultaneously supervises the reasoning process and the final answer, effectively ensuring the localization accuracy of reasoning anchors. Across 14 embodied VLM benchmarks, RoboPIN with only 4B parameters achieves an average improvement of 12\% over the strongest 7B baseline Mimo-Embodied. Overall, our work demonstrates that explicitly anchoring
  the reasoning process to visual evidence is an effective and scalable path toward enhancing the spatial reasoning capabilities of embodied VLMs. Various analytical experiments further confirm that PinCoT substantially improves the localization accuracy and cross-step identity consistency.
  Although RoboPIN already surpasses larger-scale embodied VLMs, the current work validates PinCoT only at the 4B parameter scale. In the future, scaling up data and model size is expected to further improve performance, and we plan to extend PinCoT to 3D spatial localization and long-horizon
  planning, exploring the potential of reasoning anchors in more complex physical interactions.

\bibliography{sample-base}

\clearpage
\appendix

\section{Data Construction Details}
\label{sec:appendix_construction}

\subsection{Semantic Parsing}
\label{sec:semantic_parsing}
In embodied scenarios, the environment is often cluttered, containing numerous objects and scene elements. Directly extracting entities from raw questions without structured parsing may introduce irrelevant objects, which in turn degrades the quality of the final reasoning data.

To address this issue, we design task-specific semantic parsing prompts tailored to handle a diverse range of questions. We illustrate this design using the spatial and multi-view tasks from SAT as representative examples.

For spatial understanding questions in SAT, the query typically contains explicitly relevant objects. The semantic parsing step directly extracts the primary target object alongside any auxiliary spatial references (e.g., markers). This targeted extraction reliably produces high-quality entities. In contrast, multi-view reasoning tasks in SAT, including camera rotation and camera motion, typically lack a single explicit target. Therefore, we need to identify a set of trackable objects from each image to facilitate the construction of consistent anchors across views. We provide two illustrative examples below:

\begin{tcolorbox}[
colback=gray!10!white,
colframe=gray!70!black,
arc=4pt,
boxrule=0.8pt,
left=8pt,
right=8pt,
top=8pt,
bottom=8pt,
title={Question and Structured Output},
colbacktitle=gray!75!black,
coltitle=white,
fonttitle=\bfseries,
boxed title style={
  boxrule=0pt,
  colframe=gray!75!black,
  colback=gray!75!black,
  arc=4pt
}
]
\small
I need to go to bed (near the mark 3 in the image). Which direction should I turn to face the object?

\texttt{target\_text = [bed]}\\
\texttt{marker\_refs = ["3"]}
\end{tcolorbox}

In this example, the model directly obtains the relevant entity and marker: \texttt{bed} and mark \texttt{3}, from the question.

\begin{tcolorbox}[
colback=gray!10!white,
colframe=gray!70!black,
arc=4pt,
boxrule=0.8pt,
left=8pt,
right=8pt,
top=8pt,
bottom=8pt,
title={Question and Structured Output},
colbacktitle=gray!75!black,
coltitle=white,
fonttitle=\bfseries,
boxed title style={
  boxrule=0pt,
  colframe=gray!75!black,
  colback=gray!75!black,
  arc=4pt
}
]
\small
The first image is from the beginning of the video and the second image is from the end. How did the camera rotate from the first image to the second image?

\textbf{Image 0 (first image): }

\texttt{target\_text = [couch, bookshelf, painting, doorway, table, chair, counter]}

\textbf{Image 1 (second image): }

\texttt{target\_text = [couch, doorway, table, painting, door, counter, chair]}
\end{tcolorbox}

This example corresponds to a multi-view scenario where the model parses each image individually. Since no explicit target is specified, the model selects a set of objects suitable for cross-view matching instead of extracting a single entity.

Overall, by employing task-specific prompts, the semantic parsing stage extracts highly relevant objects from the raw questions, which ultimately improves the quality of the reasoning data.

\subsection{Grounding and Reverse Verification}

Building upon the entity descriptions acquired in the semantic parsing stage, we further extract the spatial information of the corresponding objects using a pipeline of Florence-2 and SAM 2.1. Specifically, Florence-2 first generates candidate regions, from which the highest-confidence region is selected as the initial prediction. Subsequently, SAM 2.1 is applied to refine the segmentation, yielding a more stable mask representation.

From the stable mask representations generated by SAM 2.1, we extract both bounding boxes and point anchors. The point anchors are utilized to construct reasoning anchors, whereas selected bounding boxes serve to provide spatial supervision signals for $R_{\text{pin}}$ in the subsequent RFT training phase.

Although the combined Florence-2 + SAM 2.1 pipeline provides reliable grounding in general object-centric cases, it may fail for highly symbolic targets, such as text markers or abstract symbols. These targets often lack distinctive visual features, making them harder to localize accurately. For example, queries like ``mark 3'' may be mislocalized to similar regions (e.g., ``mark 0'').

To mitigate this issue and enhance overall data quality, we introduce a reverse verification step. Upon acquiring the candidate bounding boxes, we leverage Florence-2 to reinterpret the localized regions to verify their alignment with the original semantic queries. This verification is highly effective due to Florence-2's versatile detection mechanisms, which excel in OCR and region recognition tasks. Consequently, any candidate exhibiting a semantic mismatch is immediately discarded.

\begin{figure}[t]
  \centering
  \includegraphics[width=\linewidth]{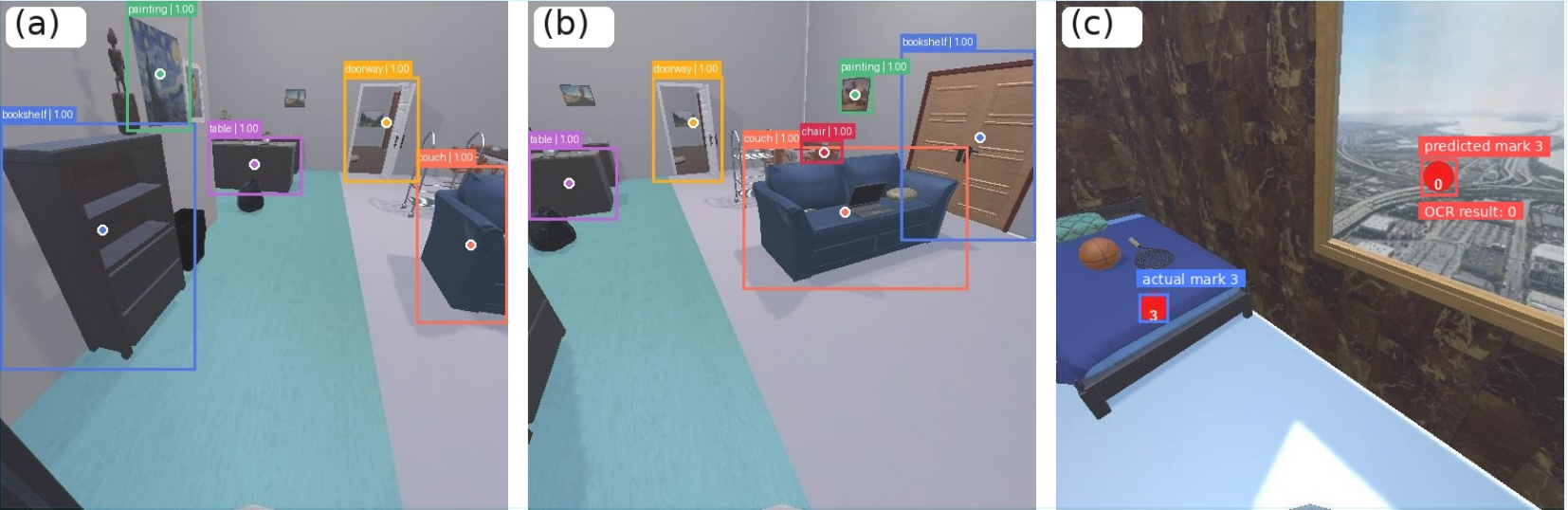}
  \caption{Representative examples of grounding and reverse verification.}
  \label{fig:appendix_grounding_reverse_verification}
\end{figure}

Figure~\ref{fig:appendix_grounding_reverse_verification} illustrates representative examples corresponding to the questions in Sec.~\ref{sec:semantic_parsing}. Panels (a) and (b) show successful grounding of anchor objects across different views in multi-view tasks. Panel (c) presents a failure case where the initial detection incorrectly localizes a symbolic marker, which is then identified as inconsistent by OCR-based reverse verification and filtered out.

\subsection{Reasoning Generation and Filtering}

Based on the entity descriptions and point anchors, we construct high-quality reasoning traces by combining them with the original questions and answers.

During this stage, object identifiers are not pre-defined but are dynamically introduced by the model during reasoning generation. The model produces structured reasoning sequences following the \pincot{} format. To further enhance the quality of the generated reasoning data, we apply strict filtering mechanisms. Specifically, we ensure that all outputs strictly adhere to the predefined structural constraints, and we discard any reasoning traces that contain corrective transition words or hesitations, such as "wait." The presence of such terms typically indicates that the model has erroneously relied on irrelevant or unsuitable objects for its analysis, leading to logical inconsistencies. Additionally, we apply an anti-leakage filter to remove sequences that explicitly rely on the ground-truth answer (e.g., phrases like “based on the provided answer”), ensuring that the reasoning remains independent rather than reverse-engineered. By enforcing these combined constraints, we effectively guarantee the structural integrity and overall reliability of the final thinking sequences. The prompt template employed to guide the generation of these reasoning sequences, along with a representative example corresponding to the questions in Sec.~\ref{sec:semantic_parsing}, is detailed as follows:

\begin{tcolorbox}[
colback=gray!10!white,
colframe=gray!70!black,
arc=4pt,
boxrule=0.8pt,
left=8pt,
right=8pt,
top=8pt,
bottom=8pt,
title={Prompt and Example for Thinking Generation},
colbacktitle=gray!75!black,
coltitle=white,
fonttitle=\bfseries,
boxed title style={
  boxrule=0pt,
  colframe=gray!75!black,
  colback=gray!75!black,
  arc=4pt
}
]
\small\ttfamily
You are given a problem description, the correct answer, and candidate object tags.\\
Your task is to generate a narrative, coherent, and fluid line of thought enclosed within think tags — that explains how the correct answer is derived.\\
Use the full tag (e.g., \texttt{<obj name="X" id="obj\_01" img\_idx="0" point="[x, y]">}) on first mention within EACH image. If the object appears in a new image, you MUST use the full tag again. For all subsequent references, use ONLY the ID text (e.g., obj\_01).\\
  - tag\_name can be obj for an object, space for a defined area, etc. \\
  - ID must follow the format "tagname\_NUMBER" (e.g., obj\_01, space\_01). \\
  - img\_idx indicates the image index, and point gives its 2D coordinates as [x, y]. \\
  - For the same object appearing across multiple views, you MUST assign and reuse the exact same ID to maintain cross-view consistency. \\
Wrap the entire reasoning in \texttt{<think>} and \texttt{</think>} tags.\\
problem:\{question\}\\
answer:\{answer\}\\
relative points:\{objects\} \\

\textbf{Here is an example:}\\
<think>I need to determine how the viewpoint changes from the first image to the second image.\\
First, I examine the <obj name="doorway" id="obj\_01" img\_idx="0" point="[760, 238]"> in the first image. In the second image, the same <obj name="doorway" id="obj\_01" img\_idx="1" point="[324, 238]"> appears much farther to the left. The significant decrease in its x-coordinate (from 760 to 324) clearly indicates that obj\_01 has shifted leftward across the frame.\\
Next, I look at the <obj name="couch" id="obj\_02" img\_idx="0" point="[926, 479]"> in the first image, where obj\_02 is partially occluded. In the second image, the same <obj name="couch" id="obj\_02" img\_idx="1" point="[623, 414]"> becomes more fully visible, suggesting that obj\_02 also shifts leftward (x-coordinate decreases from 926 to 623).

Since multiple objects show a consistent leftward shift in their coordinates, this visual evidence directly supports a rightward rotation of the camera angle.

Therefore, the camera rotated right.</think>

\end{tcolorbox}

\section{Training Datasets Details}
\label{sec:appendix_data}

Following the three-stage training pipeline described in Method section, Stage~1 (SFT) performs embodied domain adaptation using 259k samples of general embodied data, formatted as standard question-answer pairs. 
Stage~2 (CoT-SFT) uses PIN-170K as the structured reasoning corpus to learn the \pincot{} format and identity-aware reasoning mechanism, with an increased focus on spatial and multi-view reasoning tasks. 
Stage~3 (RFT) further applies reward-based fine-tuning on a 30k reward-compatible subset to strengthen reasoning ability.

Table~\ref{tab:appendix_corpus_overview} summarizes the exact sample counts of each data source in all three stages, from which the corresponding mixture ratios can be directly derived. 
These ratios are designed to align with the objectives of each stage: Stage~1 is dominated by general embodied data for domain adaptation; Stage~2 increases the proportion of structured reasoning and multi-view data; and Stage~3 adopts a more balanced subset to support stable reward-based optimization.

All point coordinates used in both the \pincot{} reasoning traces and final answers are normalized to the same 0--1000 range.

\begin{table}[t]
  \centering
  \caption{The data mixture used in the three training stages of \method{}.}
  \label{tab:appendix_corpus_overview}
  \small
  \setlength{\tabcolsep}{5pt}
  \begin{tabular}{llll}
    \toprule
    Stages & Data Type & Source & Size \\
    \midrule
    Stage 1 & Geometric Reasoning & Euclid & 30k \\
    (SFT) & General Visual Reasoning & Video-R1 & 15k \\
     & Pointing & Embodied-Point & 50k \\
     & Embodied Planning & EgoPlan & 40k \\
     & Embodied Planning & RoboVQA & 40k \\
     & Embodied Spatial Reasoning & EO-Data & 84k \\
     & Total & -- & 259k \\
    \midrule
    Stage 2 & Embodied Spatial Reasoning & EmbSpatial & 20k \\
    (CoT-SFT) & Multi-view Reasoning & SAT & 15k \\
     & Embodied Spatial Reasoning & EO-Data & 45k \\
     & Embodied Planning & RoboVQA & 5k \\
     & Pointing & Embodied-Point & 40k \\
     & Multi-view Reasoning & CrossPoint & 30k \\
     & Multi-view Reasoning & InternData-M1 & 15k \\
     & Total & -- & 170k \\
    \midrule
    Stage 3 & Embodied Spatial Reasoning & EmbSpatial & 5k \\
    (RFT) & Multi-view Reasoning & SAT & 5k \\
     & Pointing & Embodied-Point & 5k \\
     & Embodied Spatial Reasoning & EO-Data & 5k \\
     & Multi-view Reasoning & CrossPoint & 5k \\
     & Embodied Planning & RoboVQA & 5k \\
     & Total & -- & 30k \\
    \bottomrule
  \end{tabular}
\end{table}

\section{Training Configurations}
\label{sec:appendix_training}

The detailed training configurations for the three-stage post-training pipeline and the corresponding RFT settings are summarized below.

\subsection{Training Configurations Across Stages}

We summarize the hyperparameter settings of the three-stage post-training pipeline in Table~\ref{tab:appendix_training_hparams}, including the trainable part, optimization setup, hardware configuration, and approximate training time for each stage. ``Language model'' indicates that the vision encoder and multimodal connector are frozen during training. All experiments are conducted on NVIDIA A800 80GB GPUs.

\begin{table}[t]
  \centering
  \caption{Detailed configuration for each training stage of the \method{}.}
  \label{tab:appendix_training_hparams}
  \small
  \setlength{\tabcolsep}{4pt}
  \begin{tabular}{lccc}
    \toprule
    Setting & SFT & CoT-SFT & RFT \\
    \midrule
    Trainable part & Language model & Language model & Full model \\
    Per-device batch size & 6 & 4 & 2 \\
    Gradient accumulation & 2 & 2 & 8 \\
    Learning rate & $3\times10^{-6}$ & $1\times10^{-5}$ & $1\times10^{-5}$ \\
    Training epochs / steps & 3 & 2 & 3 \\
    Optimizer & AdamW & AdamW & AdamW \\
    Weight decay & -- & -- & 0.01 \\
    Warmup ratio & 0.1 & 0.1 & 0.0 \\
    LR schedule & cosine & cosine & Constant \\
    Max sequence length & 8196 & 8196 & 8196 \\
    Rollout completions & -- & -- & 4 \\
    GPU numbers & 4 & 8 & 8 \\
    Training time & 23h & 7h & 12h \\
    \bottomrule
  \end{tabular}
\end{table}

\subsection{Reward Configuration}

In this section, we specify the concrete reward weights used in the RFT stage. We adopt a composite reward with task-dependent activation based on the available supervision signals:

\begin{equation}
R = \lambda_f R_{\text{format}} + \lambda_a R_{\text{accuracy}} + \lambda_c R_{\text{consistency}} + \lambda_p R_{\text{pin}}.
\end{equation}

$R_{\text{pin}}$ is applied only to datasets with explicit target region supervision, avoiding over-constraining the model's reasoning on other tasks. In our training setup, such samples account for approximately 16\% of the total RFT data.

For this spatially supervised subset, we use $(\lambda_f, \lambda_a, \lambda_c, \lambda_p) = (0.05, 0.50, 0.20, 0.25)$. For the remaining data without explicit spatial supervision, we use $(\lambda_f, \lambda_a, \lambda_c, \lambda_p) = (0.10, 0.70, 0.20, 0.00)$.

\section{General Capability Preservation}
\label{sec:appendix_general_cap}

We evaluate whether our embodied alignment preserves general-purpose vision-language capabilities. We compare \method{} against the original Qwen3-VL-4B on representative general multimodal benchmarks. These benchmarks cover complementary aspects of multimodal capability, including perception (MME), reasoning (MMStar), and real-world understanding (RealWorldQA).

\begin{table}[h]
  \caption{Comparison between Qwen3-VL-4B and \method{} on representative general-purpose VLM benchmarks.}
  \label{tab:general_retention}
  \centering
  \begin{tabular}{lcc}
    \toprule
    Benchmark & Qwen3-VL-4B & \method{} \\
    \midrule
    MME & 87.3 & \textbf{88.8} \\
    RealWorldQA & \textbf{69.4} & 68.4 \\
    MMStar & 63.8 & \textbf{64.3} \\
    \bottomrule
  \end{tabular}
\end{table}

As shown in Table~\ref{tab:general_retention}, \method{} does not exhibit catastrophic forgetting after embodied post-training. It improves over the base model on MME and MMStar, while showing a slight drop on RealWorldQA (69.4 $\rightarrow$ 68.4). Overall, these results indicate that our training largely preserves general capabilities, with slight improvements in some cases.

\section{Prompt for Using \method{}}
\label{sec:appendix_prompt}

To ensure consistent structured reasoning behavior during deployment, we adopt the following prompting formats.

\begin{tcolorbox}[
colback=gray!10!white,
colframe=gray!70!black,
arc=4pt,
boxrule=0.8pt,
left=8pt,
right=8pt,
top=8pt,
bottom=8pt,
title={Prompt for Standard VQA},
colbacktitle=gray!75!black,
coltitle=white,
fonttitle=\bfseries,
boxed title style={
  boxrule=0pt,
  colframe=gray!75!black,
  colback=gray!75!black,
  arc=4pt
}
]
\small
Provide your thinking process between the <think> and </think> tags, and then give your final answer between the <answer> and </answer> tags.
\end{tcolorbox}
  
\begin{tcolorbox}[
colback=gray!10!white,
colframe=gray!70!black,
arc=4pt,
boxrule=0.8pt,
left=8pt,
right=8pt,
top=8pt,
bottom=8pt,
title={Prompt for Point-Grounded VQA},
colbacktitle=gray!75!black,
coltitle=white,
fonttitle=\bfseries,
boxed title style={
  boxrule=0pt,
  colframe=gray!75!black,
  colback=gray!75!black,
  arc=4pt
}
]
\small
The answer should be presented in JSON format. Provide your thinking process between the <think> and </think> tags, and then give your final answer between the <answer> and </answer> tags.
\end{tcolorbox}

\section{Additional Experiments and Analyses}
\label{sec:appendix_additional_experiments}

In this section, we provide supplementary experimental results and analyses that further support the claims made in the main paper. These include an in-depth analysis of multi-step grounding robustness, inference latency comparisons, and an extended ablation study on reasoning formats for multi-view Tasks.

\subsection{Multi-Step Grounding Analysis}
\label{sec:appendix_multistep_grounding}

To assess whether PinCoT sustains grounding accuracy over long reasoning chains, we conduct a step-depth analysis on the EmbSpatial benchmark, which contains 3,640 questions with an average of 11.4 reasoning steps. As shown in Table~\ref{tab:appendix_step_grounding}, localization predominantly occurs in steps 2–5. The strict bounding box hit rate remains high with only gradual degradation as reasoning depth increases. Crucially, unknown-identity events are negligible (0.24\%), indicating that PinCoT effectively maintains multi-step grounding without significant reference drift.

\begin{table}[htbp]
  \centering
  \caption{Grounding accuracy vs. reasoning step on EmbSpatial.}
  \label{tab:appendix_step_grounding}
  \begin{tabular}{lccccc}
    \toprule
    Localization step & 2 & 3 & 4 & 5 \\
    \midrule
    Strict bbox hit-rate (\%) & 89.40 & 87.44 & 87.23 & 81.70 \\
    \bottomrule
  \end{tabular}
\end{table}

\subsection{Inference Latency Analysis}
\label{sec:appendix_latency}

Although PinCoT introduces additional reasoning tokens during inference, we find that the overall latency impact remains modest due to the high throughput of the vLLM inference engine. We evaluate on 4,000 randomly sampled questions using a single A800 80GB GPU. As reported in Table~\ref{tab:appendix_latency}, RoboPIN emits approximately twice the number of tokens per sample compared to RoboBrain2.0 (411.67 vs. 212.18), yet the per-sample time increases only moderately (0.281 s vs. 0.212 s). This is attributed to RoboPIN's substantially higher token throughput (1462.73 tok/s vs. 1000.70 tok/s).

\begin{table}[htbp]
  \centering
  \caption{Inference efficiency comparison (vLLM, A800 80GB).}
  \label{tab:appendix_latency}
  \begin{tabular}{lccc}
    \toprule
    Model & Tokens/sample & Time/sample (s) & Tokens/sec \\
    \midrule
    RoboPIN & 411.67 & 0.281 & 1462.73 \\
    RoboBrain2.0 & 212.18 & 0.212 & 1000.70 \\
    \bottomrule
  \end{tabular}
\end{table}

\subsection{Extended Reasoning Format Ablation for Multi-View Tasks}
\label{sec:appendix_anchor_ablation}

To further investigate the effectiveness of point-based anchors in multi-view scenarios, we conduct an extended ablation study on the CrossPoint-30K dataset under the same SFT setting and training recipe. While the main paper presents multi-view experimental results, this ablation provides additional granularity by comparing different reasoning formats. We compare PinCoT against two alternatives: (i) \textbf{Coord CoT w/ ID}, which uses persistent bounding box anchors with IDs, and (ii) \textbf{Coord CoT}, which uses coordinate-based reasoning without explicit persistent IDs. As shown in Table~\ref{tab:appendix_anchor_format}, PinCoT achieves the best accuracy (71.20\%), outperforming Coord CoT w/ ID (68.50\%) and Coord CoT (67.70\%). This confirms that PinCoT better support cross-view consistency compared to .

\begin{table}[htbp]
  \centering
  \caption{reasoning-format ablation on CrossPoint-30K.}
  \label{tab:appendix_anchor_format}
  \begin{tabular}{lcc}
    \toprule
    Reasoning format & Accuracy (\%) \\
    \midrule
    PinCoT & \textbf{71.20} \\
    Coord CoT w/ ID & 68.50 \\
    Coord CoT & 67.70 \\
    \bottomrule
  \end{tabular}
\end{table}

\section{Real-World Experiment Setup}
\label{sec:appendix_real_world}

Our real-world desktop manipulation tasks are evaluated with an xArm 6 robotic arm. The setup includes an Intel RealSense L515 LiDAR camera, a wrist-mounted Intel RealSense D435 depth camera, and a force-torque sensor on the xArm to enable compliance control, thereby improving interaction with the environment. Both cameras operate at a resolution of 640$\times$480. A computer running Ubuntu 24.04 and equipped with an NVIDIA RTX 5090 is directly connected to the robotic arm and cameras to execute low-level control policies. 

For physical execution, the target manipulation points are directly predicted by models and projected into 3D Cartesian space using the corresponding depth data and camera intrinsics. A motion planner is then employed to generate collision-free paths, guiding the robotic arm's end-effector to the predicted targets for real-world execution.

\end{document}